\documentclass[runningheads]{llncs}

\usepackage{xcolor}

\usepackage[T1]{fontenc}

\usepackage{amssymb}
\usepackage{graphicx}
\usepackage{bm}
\usepackage{placeins}
\begin{document}
\title{Few-Part-Shot Font Generation}
%
%
\author{
    Masaki Akiba\inst{1} 
    \and Shumpei Takezaki\inst{1}
    \and Daichi Haraguchi\inst{2}
    \and Seiichi Uchida\inst{1}
}
\authorrunning{M. Akiba et al.}
%
\institute{
    Kyushu University, Fukuoka, Japan\\
    \email{\{masaki.akiba, shumpei.takezaki\}@human.ait.kyushu-u.ac.jp}\\
    \email{uchida@ait.kyushu-u.ac.jp}
    \and
    CyberAgent, Tokyo, Japan\\
    \email{haraguchi\_daichi\_xa@cyberagent.co.jp}
}
\maketitle              
\begin{abstract}
This paper proposes a novel model of few-part-shot font generation, which designs an entire font based on a set of partial design elements, i.e., partial shapes. Unlike conventional few-shot font generation, which requires entire character shapes for a couple of character classes, our approach only needs partial shapes as input. The proposed model not only improves the efficiency of font creation but also provides insights into how partial design details influence the entire structure of the individual characters.

\end{abstract}
%
%

\section{Introduction\label{sec:intro}}
New fonts are created and published almost daily through the considerable efforts of typeface designers. Even for the basic uppercase letters (``A'' to ``Z''), they must define 26 distinct character shapes. Furthermore, each shape must be carefully refined to ensure the font's style remains unique and consistent across all letters.
\par 

Various software tools have been developed to assist in font design. For example, {\tt FontLab}\footnote{\tt https://www.fontlab.com} and {\tt Glyphs}\footnote{\tt https://glyphsapp.com} are widely used font editors, enabling designers to create typefaces from scratch or select a reference font and gradually refine it to achieve their desired style. These tools now offer advanced features for creating so-called {\em variable fonts}, in which attributes such as slant and stroke width can be controlled by parameters. While these tools reduce manual effort, designers must still make numerous fine adjustments to keep the font style unique and consistent across the entire set of characters.
\par 

\begin{figure}
\centering
\includegraphics[width=\textwidth]{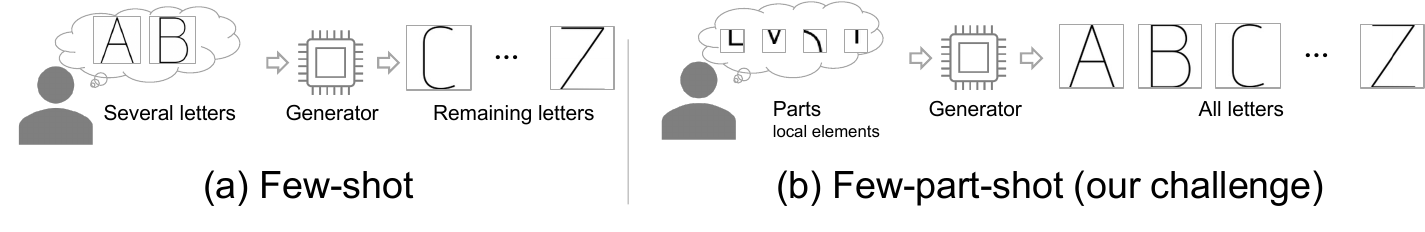}\\[-3mm]
\caption{(a)~Conventional few-shot font generation. (b)~Our challenge of few-part-shot font generation. We aim to clarify the capabilities and limitations of (b) by benchmarking it against (a). \label{fig:purpose}}

\bigskip
\includegraphics[width=0.9\linewidth]{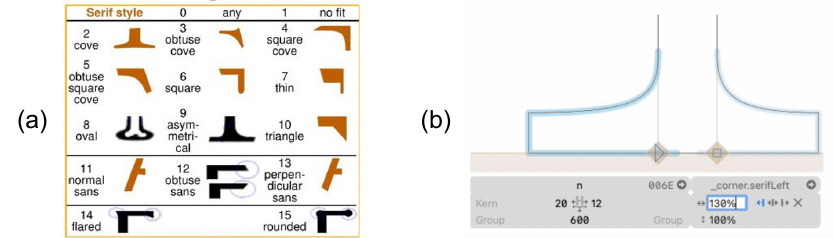}
\caption{(a)~{\tt PANOSE} Font Classification Chart (in part) (source: https://archive.org/details/Panose-chart). (b)~A snapshot of the part-driven font design function in {\tt Glyphs} (source: https://glyphsapp.com/learn/reusing-shapes-corner-components).\label{fig:panose}}

\end{figure}

Recently, generative models --- such as Generative Adversarial Networks (GANs)\cite{goodfellow2014generative} and diffusion models\cite{NEURIPS2020_DDPM} --- have been applied to font design, giving rise to a semi-automatic framework called {\em few-shot font generation}. It generates all characters in the alphabet using only one or a few fully designed reference characters\footnote{If a framework assumes one reference character, it is called one-shot font generation.}. For instance, as illustrated in Fig.~\ref{fig:purpose}(a), by providing the character images of ``A'' and ``B'' in a certain style, the remaining letters ``C'' through ``Z'' can be generated in that same style. Naturally, this framework holds great potential to reduce the effort required in font design.
\par

This paper challenges a new framework called few-{\em part}-shot font generation, which designs an entire font from a small set of user-specified parts. {\em Parts} are defined as representative local shapes capturing stroke thickness, terminations, corners, curves, intersections, and decorative elements. Parts play important roles in specifying font styles. The PANOSE chart~\footnote{\url{https://archive.org/details/Panose-chart}} in Fig.~\ref{fig:panose}(a) is a well-known font style classification scheme and uses part shapes. The above font-designing software tools ({\tt FontLab} and {\tt Glyphs}) already offer functionalities of part-oriented design, as shown in Fig.~\ref{fig:panose}(b).\par

Fig.~\ref{fig:purpose}(b) illustrates the idea of few-part-shot font generation. 
This idea stems from the observation that the same or similar parts appear in multiple characters.
For instance, defining a vertical stroke end as a part facilitates the design of characters such as ``I,'' ``K,'' or ``P,'' all of which share a similar vertical stroke pattern. Consequently, even a small set of well-chosen parts can effectively represent the majority of local shapes in the alphabet.

Despite the anticipated difficulties, the challenge of this new few-part-shot font generation offers three novel insights.
First, from a practical perspective, it offers a new degree of flexibility in font creation. As illustrated in Fig.~\ref{fig:purpose}(a), the existing framework requires at least one fully designed reference character image; however, in an extreme example where the single reference character is ``O,'' no information about straight lines would be provided, potentially hindering the generation of other characters. By contrast, our new framework imposes no requirement to complete the design of an entire character image; users can simply supply arbitrary parts, particularly those shapes deemed critical for the desired font style.\par

\begin{figure}[t]
    \centering
    \includegraphics[width=0.5\textwidth]{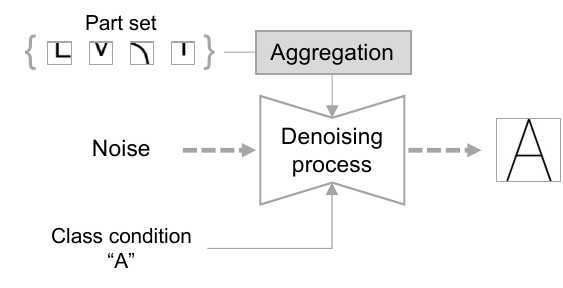}\\[-3mm]
    \caption{Overview of the proposed model. 
    The part set is aggregated into a single style feature vector. The diffusion model then generates the target font image conditioned on this vector and the character class.} \label{fig:overview}
\end{figure}

Second, there is a scientific motivation in that this framework can reveal the extent to which a limited set of parts correlates with an entire alphabet. If a small collection of parts can reproduce most character shapes, those parts effectively function as a compressed representation of the overall shapes in the alphabet.\footnote{
it is well-known that parts capture character ``classes,''  through various part-based classification techniques such as bag-of-visual-words\cite{wang2013part,uchida2010part}. In contrast, our focus here is on how well parts capture complete character ``shapes.''} To our knowledge, no studies have addressed the question of how parts can faithfully reconstruct entire characters, suggesting that our attempt could be a new starting point for font research.\par

Third and finally, it is also crucial to clarify the limitations of this framework. Few-part-shot generation cannot, in principle, perfectly reconstruct certain global characteristics from only local parts. A simple illustration is a rectangle: from the four 90-degree corners of a rectangle, we cannot reconstruct the original rectangle because the corners do not fix its aspect ratio. Applying this to fonts, even if all parts come from a ``condensed''  font (i.e., a narrow font), there is no guarantee that the original condensed appearance can be recovered precisely. Likewise, overall features such as oblique angles may lie outside the representational reach of a small set of parts. Despite these potential shortcomings, our experiments will demonstrate that few-part-shot generation often produces surprisingly reasonable results in practice. Moreover, future work can focus on a complementary combination of the conventional entire-shape-based and our part-based generation frameworks.
\par

In our practical implementation of few-part-shot font generation, we employ a diffusion model that is conditioned on a set of parts. This setup ensures that the generated character images reflect the stylistic cues embedded in those parts. Fig.~\ref{fig:overview} presents a simplified overview of the proposed model. To handle a variable number of user-defined parts, we use DeepSets~\cite{NIPS2017_f22e4747}, which transforms each part into a feature vector and then aggregates all such vectors into a single, unified representation. This design allows our model to integrate any number of parts without modifying the network architecture, thereby keeping the style feature vector consistent regardless of the set’s size.\par

\section{Related Work}
\subsection{Few-shot font generation}
A common approach to example-based font generation is few-shot font generation, which aims to generate complete fonts by extracting style features from a small number of example characters.
In the early stages of the research, EMD~\cite{zhang2018separating} and AIGS-Net~\cite{Gao2019Artistic} extracted both content and style features directly from whole character images to synthesize the target font. 

In recent years, many studies have shifted their focus to the generation of fonts for complex writing systems such as Chinese and Hangul, which contain component (radical) structures~\cite{kong2022look,park2021few,park2021mxfont}. To address this complexity, methods that leverage component-based content and style feature extraction have been proposed. In the GAN-based approaches, LF-Font~\cite{park2021few} and DM-Font~\cite{cha2020dmfont} uses component labels for style extraction, and CG-GAN~\cite{kong2022look} utilizes weak component labels in the discriminator for one-shot font generation. FS-Font~\cite{Tang_2022_CVPR} captures fine-grained local styles from the whole image via cross-attention.
Furthermore, Diff-Font~\cite{he2024diff}, a diffusion-based approach, incorporates stroke-level information to enhance font generation quality. These component-based approaches have achieved high accuracy by decomposing complex characters, such as Chinese characters, into their components, such as radicals. 

Thanks to the development of the diffusion model, several methods that do not rely on component information have achieved state-of-the-art performance. For example, MSD-Font~\cite{fu2024generate} employs a three-stage reverse diffusion process. FontDiffuser~\cite{yang2024fontdiffuser}, on the other hand, achieves one-shot font generation by utilizing multi-scale feature maps while ensuring font style consistency through its Style Contrastive Refinement module, which aligns the font style features from generated font and reference font image.

This paper focuses on Latin alphabets, which do not have component structures and proposes a novel approach that extracts font style features from parts (i.e., local shapes) of characters to generate complete font images. This perspective can be seen as further refining the granularity of feature extraction compared to conventional component-based methods. 

\subsection{Part-based analysis of fonts}

The analysis of fonts based on individual parts is not a common topic in computer science but was first introduced by Ueda et al.~\cite{ueda2021parts}.
Their study focused on not analyzing the global shape of a font but examining individual parts, aiming to identify which parts contribute to specific impressions, such as ``legible,'' ``elegant,'' or ``comic-like.''  As an extension of the first study, they further investigated the association between specific parts and impressions~\cite{ueda2022font}. They hypothesized that the combinations of parts, rather than individual parts alone, play an important role in shaping font impressions. To examine this hypothesis, they employed a self-attention mechanism\cite{NIPS2017_3f5ee243} to capture dependencies between parts and analyzed their joint influence on impressions. Consequently, those trials emphasize how local parts of font are important.

\section{Few-Part-Shot Font Generation}

Our proposed few-part-shot font generation method has a relatively simple architecture. Specifically, it consists of two main stages: (1) font style feature extraction from the given set of parts and (2) font image generation using a diffusion model. As will be shown in the experiments, even this rather simple architecture still successfully generates font images from the limited information contained in the parts. This finding not only validates the sufficiency of our proposed method but also highlights that the overall design of a character strongly depends on its local shapes --- in other words, knowing the local shapes provides a clear pathway to reconstructing the entire character.

\subsection{Font style extraction from parts}
We now describe how to pre-train the model to extract a consistent font style representation from an arbitrary number of parts. To learn the mapping from a given set of parts to a style feature vector, we employ {DeepSets~\cite{NIPS2017_f22e4747}}. Furthermore, we use contrastive learning to ensure that the style feature vector can capture subtle variations in font style while remaining invariant to character classes (e.g., ``A'' and ``B'' of the same font should share the same style feature).
\par

\begin{figure}[t]
\includegraphics[width=\textwidth]{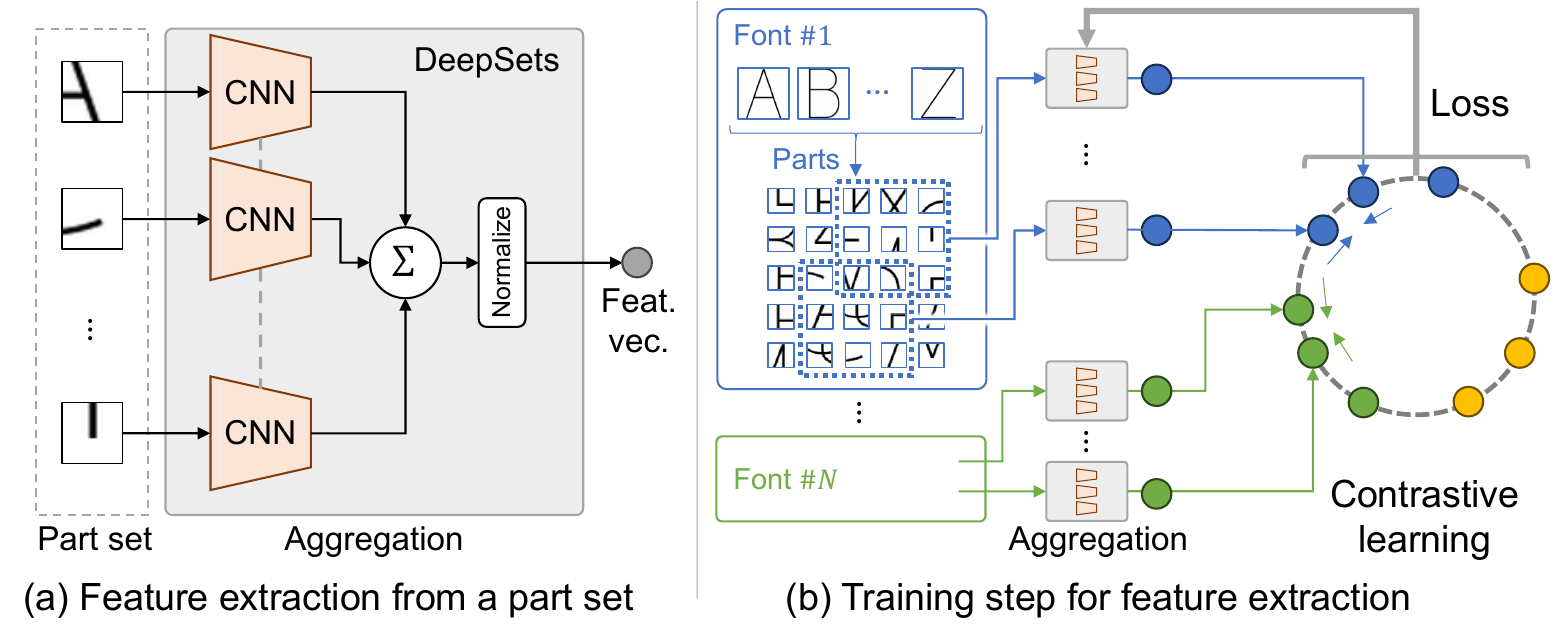}
\caption{Style feature extraction from a set of parts using DeepSets and contrastive learning. 
(a) Each part in the part set is encoded into a feature vector using a CNN. The resulting vectors are summed and then normalized to produce a single aggregated style vector. (b) The feature extractor is pretrained using contrastive learning, which pulls together part sets from the same font and pushes apart those from different fonts.
\label{fig:stylefeatextraction}}
\end{figure}

\subsubsection{Aggregating a set of parts using DeepSets}
DeepSets is a type of neural network model designed to handle sets with an arbitrary number of elements. As depicted in Fig.~\ref{fig:stylefeatextraction}(a), DeepSets consists of two components: (i) a network that transforms each element of the set into a vector, and (ii) an operator that aggregates the resulting vectors. In our case, each part is given as a small image (i.e., a patch), so we use a CNN for the transformation. The aggregation is simply a standard vector addition. After the addition, the resultant vector is normalized to have a unit norm, yielding the aggregated style feature vector. This normalization is intended to reduce the dependence of the feature vector on the number of parts in the set.
\par

\subsubsection{Contrastive learning}

The style feature vector obtained in DeepSets is refined via contrastive learning\cite{journals/corr/abs-1807-03748}, as illustrated in Fig.~\ref{fig:stylefeatextraction}(b). Contrastive learning encourages feature vectors with the same label to be pulled closer together, while pushing those with different labels apart. We apply this mechanism to sets of parts originating from the same font, forcing their style vectors to be similar. We also apply it to sets from different fonts, driving their style vectors apart and enabling the model to learn a stable representation of the font style. Through this procedure, the internal CNN of DeepSets is trained to produce style vectors that depend as little as possible on the specific parts chosen. Furthermore, by training with part sets of varying sizes, we ensure that the resulting style feature vector remains robust regardless of the size of the set.

\subsection{Font generation with the style feature from a part set}

As shown in Fig.~\ref{fig:overview}, we employ a standard conditional diffusion model to generate complete font images from the style feature vectors produced by DeepSets. Concretely, we train a U-Net that progressively transforms random noise into an entire character image (rather than just a partial one), conditioning on three key inputs: (1) the character class label (e.g., ``A''---``Z''), (2) the style feature vector extracted from the parts, and (3) the diffusion process timestep, which is omitted in Fig.~\ref{fig:overview} for simplicity.
\par

During training, we integrate both the character label (converted into an embedding) and the style feature vector (obtained from a user-provided set of parts via DeepSets) into the U-Net through cross-attention layers. This design allows the network to adapt its denoising process to the target character and the specified font style. In addition, we randomly vary the number of parts in each set, enabling the model to handle different amounts of local shape information. As a result, at inference time, the U-Net can generate high-quality font images for any character using only the chosen style parts. 

\section{Experimental Setup}
\subsection{Dataset}

In this study, we used 26 uppercase letters (``A''--``Z'') from the Google Fonts\footnote{\tt https://fonts.google.com} dataset for model training and evaluation. The collected dataset consists of a total of 1,804 font families and 6,796 fonts. We divided them into three mutually exclusive font-family subsets to minimize the similarity between training and testing fonts. Specifically, 5,326 fonts from 1,443 font families were used for training. 746 fonts from 181 font families were used as the validation set for hyperparameter tuning and early stopping. The remaining 724 fonts from 180 font families are used for quantitative and qualitative evaluations.
Each character in the font sets is rendered to be a $128\times 128$-pixel bitmap image.

\subsection{Preparation of part set}

To train a neural network for few-part-shot font generation, we need a dataset containing only the essential parts of each font. It would be ideal if we have a public dataset of parts prepared by professional designers. However, no such public dataset exists at present. While it is possible to extract parts randomly from font images, this naive approach lacks objectivity and reproducibility, making it unsuitable for constructing a reliable dataset.

To address this issue, we propose an objective method that automatically extracts $K>0$ representative parts from each font using the Scale-Invariant Feature Transform (SIFT)~\cite{journals/ijcv/Lowe04} and $K$-medoids clustering. This method enables consistent selection of characteristic parts within a font.
\par

Extracting only the “meaningful” parts of a font is inherently challenging, since the importance of each part can vary depending on the designer’s intention and the perception of font users. Nevertheless, we assume three key properties of font design: (1)~local shapes such as serifs, stroke terminals, intersections, and corners are widely regarded as representative stylistic elements, as shown in Fig.~\ref{fig:panose};
(2)~designers typically do not reuse exactly identical parts multiple times;
and (3)~purely blank regions (i.e., background-only patches) are not considered as valid design components.

Based on these assumptions, we {\em simulate} the selection of meaningful parts by using SIFT to detect key points along the font strokes and generate corresponding feature descriptors, which are then clustered via $K$-medoids.
We finally extract $K$ parts (i.e., image patches)  corresponding to the medoid key points. The part size is a hyper-parameter to be fixed before training; we use $32\times 32$ or $64\times 64$ for $128\times 128$ font images  This procedure offers an objective and reproducible way to extract representative parts from any given font. In our experiments, we set $K$ at 26, resulting in 26 representative parts per font, although it does NOT aim to extract a single part from each of the 26 alphabets.\par

\begin{figure}[t]
\centering
\includegraphics[width=0.7\textwidth]{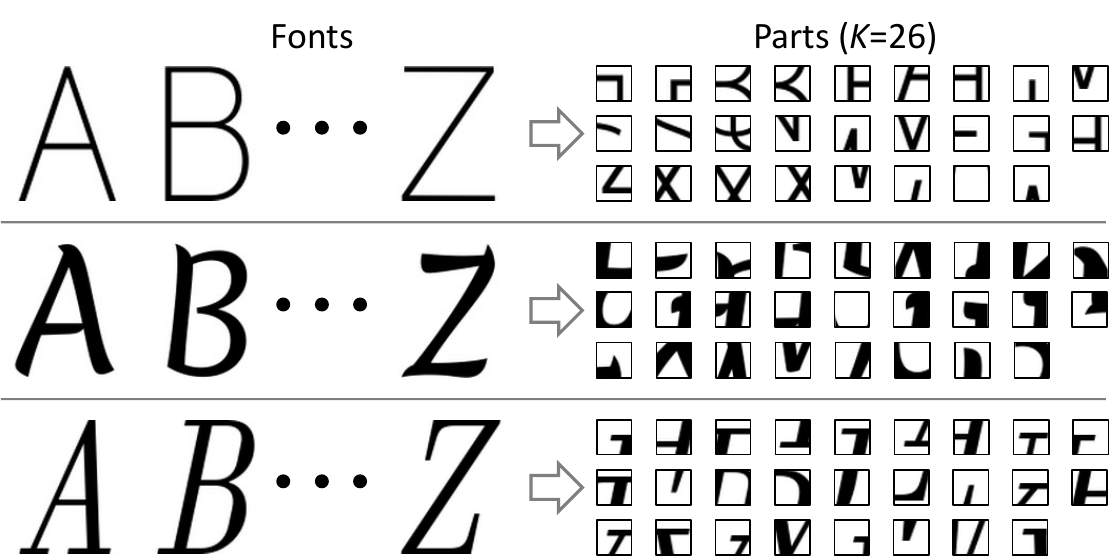}
\caption{Examples of three different fonts (left) and their corresponding set of $K=26$ parts (right) as $32\times 32$-pixel images. The extracted parts include corners, strokes, and terminations that characterize the font style.} \label{fig:part-exam}
\end{figure}

Fig.~\ref{fig:part-exam} shows results of automatically extracting 26 representative parts from each font. 
Their size is $32\times 32$ pixels. Despite the variations in overall font design (e.g., stroke thickness, curvature, or decorative elements), the method successfully identifies salient local shapes that capture key stylistic features of the font. This demonstrates how the proposed pipeline is able to distill a font’s characteristic elements into a manageable collection of parts.

\subsection{Implementation details}

\subsubsection{Font Style Extractor}
A convolutional neural network (CNN) is employed as a feature extractor for individual parts (i.e., tiny image patches) in the set, producing a 256-dimensional output vector.
During training the CNN in the DeepSets architecture of Fig.~\ref{fig:stylefeatextraction}(a), the number of parts used is randomly varied between 1 and 8. The mini-batch size is set to 64. For contrastive learning, the number of negative samples is set to 63, and the temperature parameter is fixed at 1.
The model is optimized using the Adam optimizer with a learning rate of $10^{-4}$. Additionally, early stopping is applied, terminating training if the validation loss does not improve for 500 epochs.\par

\subsubsection{Diffusion Model}
We employ a U-Net\cite{kang2024grif} for noise estimation in the diffusion process. The diffusion model operates with 1,000 timesteps. The character classes used in training consist of 26 uppercase letters `A'--`Z,' and each character class condition (represented as a 26-dimensional one-hot vector) is embedded into a 256-dimensional vector. While training the denoising model, the CNN in DeepSets is also updated. Again, the number of parts during training is randomly varied between 1 and 8. The mini-batch size is set to 64. The model is optimized using the Adam optimizer with a learning rate of $10^{-4}$. Training is conducted for approximately 200k iterations. During evaluation, the generated results are obtained after 200 sampling steps.

\section{Experimental Analysis of Few-Part-Shot Font Generation}
\subsection{Qualitative analysis}
\begin{figure}[t]
    \centering
    \includegraphics[width=\textwidth]{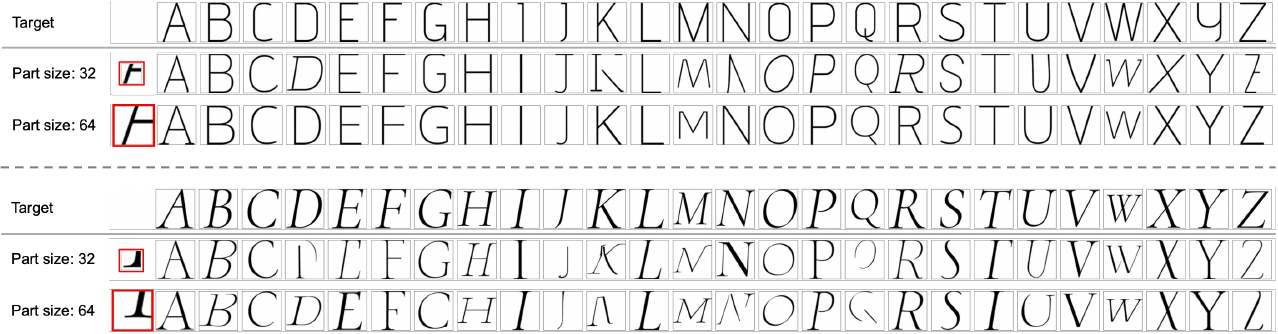}\\[-3mm]
    \caption{Results of one-part-shot font generation. Two different part sizes ($32\times 32$ and $64\times 64$) were used.}
    \label{fig:partsize}
\end{figure}
\subsubsection{One-part-shot generation for observing the basic performance}

As the simplest demonstration of our method's behavior, Fig.~\ref{fig:partsize} shows the results of {\em one}-part-shot font generation with $K=1$. The part used in the upper example is manually selected from near the horizontal stroke endpoint of the letter ``A'' in the target font, and the lower example uses a part from near another stroke endpoint of the same letter ``A.'' Even though each part is just a small fragment, it carries information about stroke thickness and curvature (or straightness). The generated alphabets reflect these features, indicating that a single part is surprisingly rich in style-related information. Although the results do not exactly match the target fonts, one can see how effectively the local shape properties are generalized into the entire alphabet.

\begin{figure}[t!]
\includegraphics[width=\textwidth]{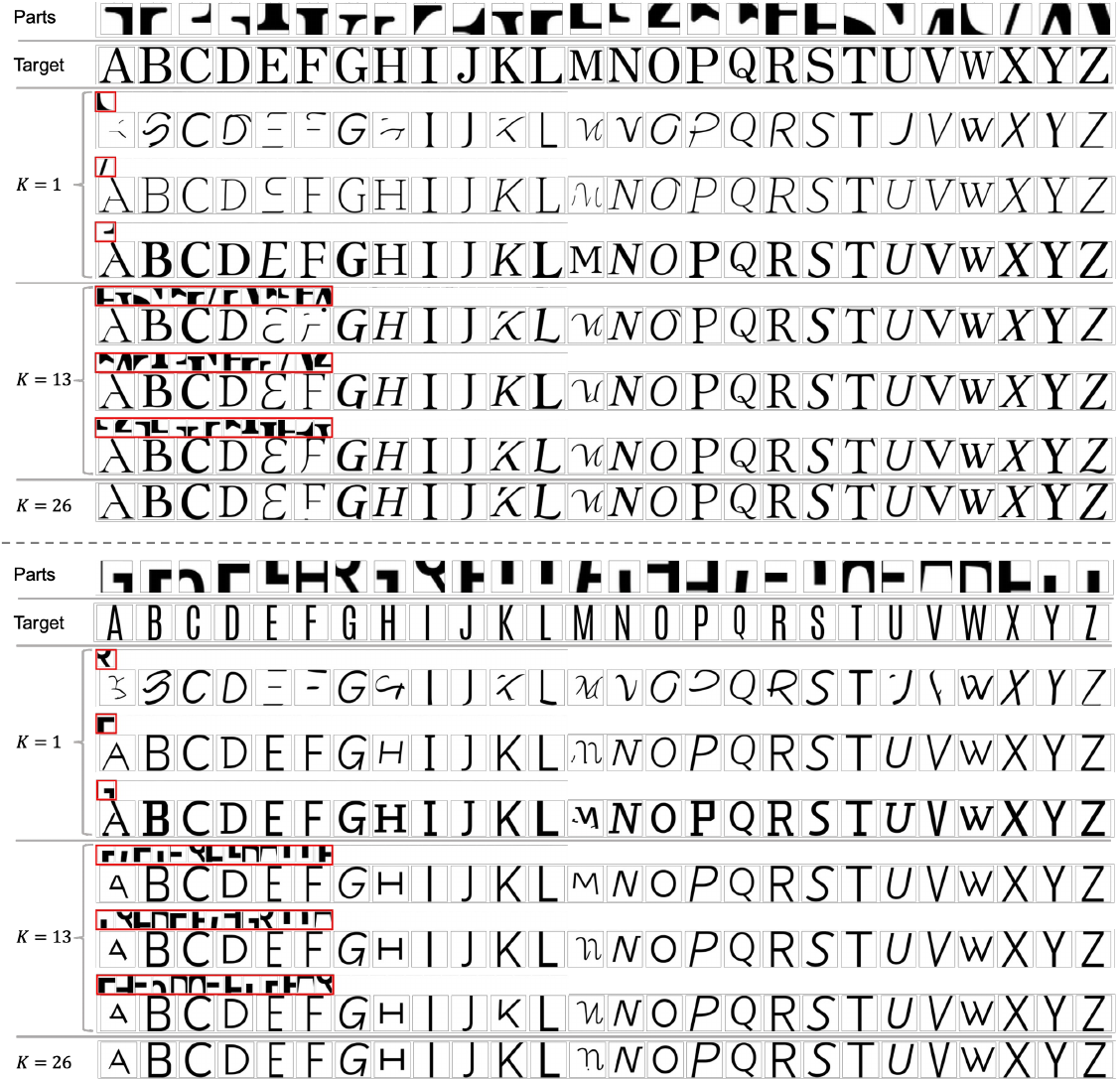}\\[-3mm]
\caption{Results of our few-part-shot font generation method under different part sets and the number of parts. The parts ($32\times 32$) used for generating the displayed result are highlighted with a red frame. For $K=26$, all of the 26 parts shown at the top were used.}\label{fig:part-influence}
\end{figure}

\subsubsection{Effect of part size}

Fig.~\ref{fig:partsize} also shows the difference when using two part sizes ($32\times 32$ and $64\times 64$) for the same region in each font. Naturally, a larger part provides more global cues. In the upper row, the 32\(\times\)32 part captures most features of the target font, yet some letters (like ``D'' or ``R'') appear in an italicized style that was not intended. Because italicization is a more global property, relying solely on a smaller local part can lead to ambiguity. In contrast, using a 64\(\times\)64 part stabilizes the style, suggesting that even moderately enlarged parts can help preserve the intended global design.
\par

In the lower row of the figure, most generated letters already include serif details even at a 32\(\times\)32 part size. This implies that local decorations, such as serifs, can be captured accurately from relatively small parts. However, in cases where the target font is italic, certain letters still fail to reflect italicization at the smaller part size. When the part size is increased to 64\(\times\)64, more letters correctly exhibit the italic style. Nevertheless, achieving perfect global consistency remains challenging with local parts alone, highlighting a fundamental limitation of few-part-shot font generation.

\subsubsection{Effect of number of parts and their variation}

Fig.~\ref{fig:part-influence} shows the results of our few-part-shot font generation when varying both the individual parts and the number of parts. For each of the two font examples shown, we extract 26 distinct parts from the target font. Note that the lower example in Fig.~\ref{fig:part-influence} features a ``condensed'' font with a notably narrow width, providing an interesting challenge for our method. By comparing the two examples, we can examine how well our approach generalizes to different styles with varying constraints.

One-part-shot generation ($K=1$) often produces images that exhibit instability or inconsistency across `A' to `Z,' depending on which single part is chosen. Nonetheless, as demonstrated in Fig.~\ref{fig:partsize}, selecting a particularly informative part (such as the third one-part-shot in the upper example) can yield generated results remarkably close to the target style. This suggests that even a single part can carry substantial style information if chosen wisely.

Increasing the number of parts $K$ raises the likelihood of a stable, coherent output. At $K=13,$ the generated characters become more consistent across all letters, as illustrated by the fully serifed style in the upper example and the consistent sans-serif style in the lower example. In the upper example, certain letters (e.g., `D' and `P') appear nearly indistinguishable from their counterparts in the target font. A crucial observation here is that the results become more robust to the specific choice of parts: different subsets of 13 parts often yield very similar character images. This phenomenon points to a strong correlation between local shapes and overall font structure. When $K=26,$ the generated outputs closely resemble those obtained at $K=13,$ indicating that even a larger set of parts may not entirely capture all aspects of the global style. We will revisit this point in a later quantitative evaluation.

In Section~\ref{sec:intro}, we discussed the fundamental limitation of relying on parts to describe an entire shape --- the rectangle example shows that merely knowing the four 90-degree corners does not guarantee recovery of its aspect ratio. This highlights the inherent ambiguity when reconstructing a global shape from local elements. In the lower example of Fig.~\ref{fig:part-influence}, the generated letters fail to maintain the narrow proportions of the condensed style, suggesting that the parts alone cannot fully capture certain global traits. Nonetheless, it is notable that much of the local styling (such as serif or sans-serif features) is still successfully reproduced.

Despite this fundamental limitation, many of the generated fonts in Fig.~\ref{fig:part-influence} exhibit a high degree of similarity to their respective targets, underscoring the promise of few-part-shot font generation. In practical settings, a designer could augment these part-based inputs with a small number of full-character references or additional global constraints, thereby refining the overall aesthetic. Such a hybrid approach would not only help overcome the current limitations of part-only generation but could also improve the fidelity of existing full-character methods by incorporating more precise local structures. This combination of global and local shape information suggests a valuable direction for future research on font generation.

\subsection{Quantitative analysis}

\subsubsection{Evaluation metrics}~\label{sec:eval_metrics}

We employ three evaluation metrics: L1 distance, SSIM, and local L1 distance. The L1 distance measures the mean absolute pixel-wise difference, thereby capturing the pixel-level error between the target and generated images. SSIM evaluates image quality by considering luminance, contrast, and structural information, rather than direct pixel-wise differences.
\par

In addition to these two metrics, we propose the local L1 distance, which focuses on the local parts of font images. Specifically, we use SIFT to detect keypoints on the target images and extract corresponding local parts. We then locate the same regions in the generated images and measure the L1 distance between the target and generated patches. This approach allows us to assess how accurately local font structures are reproduced.

\subsubsection{Quantitative evaluation by part size and number of parts}

\begin{figure}[t]
    \centering
    \includegraphics[width=1\linewidth]{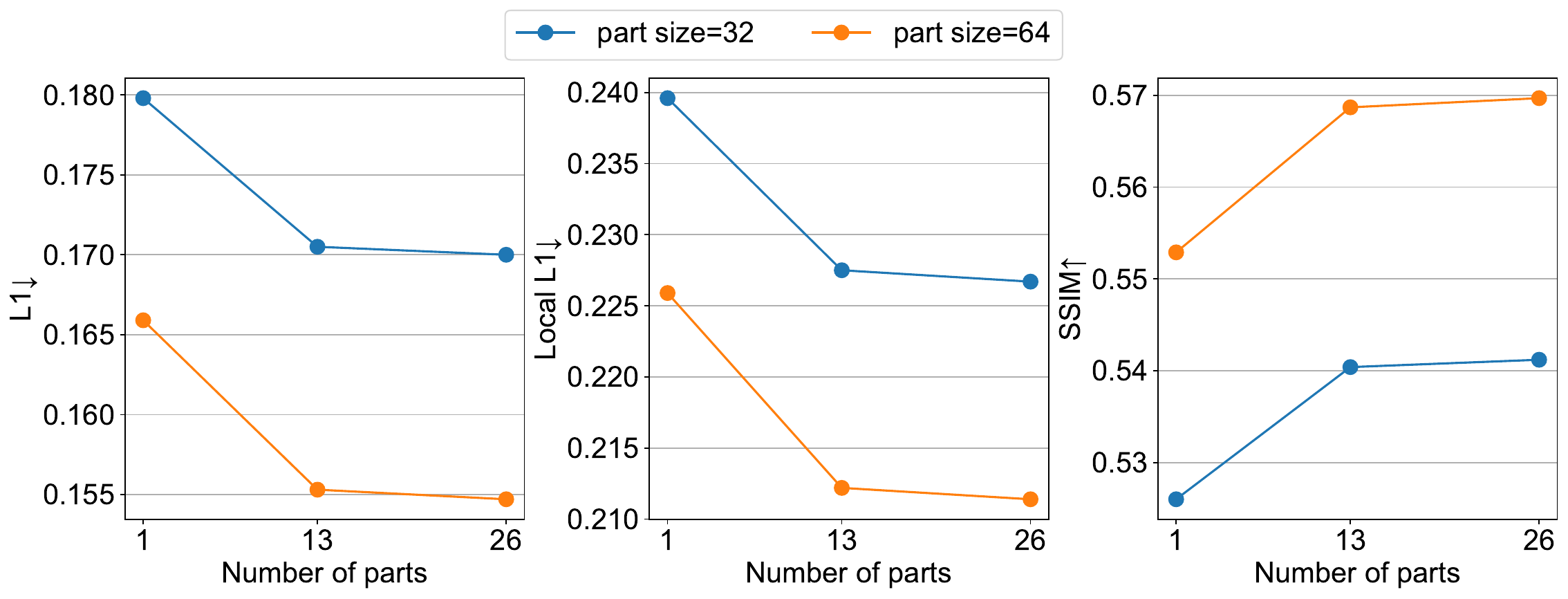}
    \caption{Quantitative performance of few-part-shot font generation as a function of part size (32 vs.\ 64) and number of parts ($K=1,\,13,\,26$). The three panels respectively show L1 distance (left), local L1 distance (middle), and SSIM (right).\label{fig:ours_eval}}
\end{figure}

Fig.~\ref{fig:ours_eval} presents the quantitative results of few-part-shot font generation when varying both the part size and the number of parts. Focusing first on part size, we observe that larger parts yield better performance in all three metrics. This result suggests that increasing the part size enriches the style information contained within a single part, thereby improving generation quality.
\par

Regarding the number of parts, increasing $K$ from 1 to 13 significantly improves all three metrics. However, the gain from 13 to 26 is relatively modest, indicating a form of diminishing marginal returns that often arises in set-based optimization problems. While adding more parts can capture additional local details, it does not necessarily guarantee a dramatic increase in global fidelity. This phenomenon also highlights a limitation: there appears to be an upper bound to how much additional style information can be conveyed by simply supplying more parts.\footnote{This phenomenon, where additional inputs yield progressively smaller benefits, is commonly referred to as {\em diminishing marginal returns.} It is observed in various fields such as economics and resource allocation.}
 (From a positive viewpoint, human designers do not need to give more and more parts to improve the quality of the generated font images.)

Interestingly, when the part size is 64, using 26 parts of size 32 still cannot surpass the accuracy of a single 64\(\times\)64 part in terms of L1 distance and SSIM. This finding reaffirms the importance of part size for capturing critical style cues. On the other hand, if we focus on local L1 distance, using 26 parts at a size of 32 yields performance nearly comparable to that of a single 64\(\times\)64 part. Hence, a sufficient number of smaller parts can effectively represent local structures, even if they do not convey as much global information as a larger part might.

\subsection{Comparative experiments}
\subsubsection{Evaluation metrics}
In addition to the evaluation metrics described in Section~\ref{sec:eval_metrics}, 
we use MSSIM, StyleFID, and ContentFID~\cite{9423371}. MSSIM is similar to SSIM and evaluates image quality by considering luminance, contrast, and structural information rather than direct pixel-wise differences. 
FID assesses the difference between the feature distributions of the generated and real images using the Fréchet distance. StyleFID is computed using feature vectors from a CNN-based font style classifier, while ContentFID is derived from a CNN-based character class classifier.

\subsubsection{Comparative methods}

One major difficulty in evaluating our approach is the lack of suitable baselines. 
To the best of our knowledge, this is the first attempt to generate fonts directly from a set of parts, so no existing models allow for a direct comparison. Moreover, many recent few-shot font generation methods focus on Chinese characters, leveraging radical (sub-character) structures~\cite{kong2022look,park2021few,park2021mxfont}. 
Since our current work targets the more decorative Latin alphabet, we cannot fairly compare with such radical-based approaches.
\par
To address this challenge, we adopt a {\em one}-shot font generation setting for fair comparison and prepare three comparative methods:
\begin{itemize}
    \item First, we use FontDiffuser~\cite{yang2024fontdiffuser} -- a state-of-the-art model for one-shot font generation without assuming radical structures -- to generate the Latin alphabet. FontDiffuser achieves high performance in one-shot font generation through a carefully designed training scheme, including two-stage learning and style refinement.
    \item Second, we also create another baseline model that employs the same diffusion backbone as ours but replaces our style feature extraction module with the encoder from FANnet~\cite{Roy_2020_CVPR}. This baseline is necessary because FontDiffuser and our proposed method differ in their underlying diffusion architectures, complicating a direct comparison. Kondo \textit{et al.}~\cite{kondo2024font} similarly used FANnet to extract font style features, combining it with a diffusion model for font generation. We thus regard it as a reasonable one-shot baseline, which we call \textbf{FANnet + DM}. Like FontDiffuser, this baseline processes a single full-character image rather than a small part.
    \item Third, we modify the proposed method to be evaluated in the same one-shot framework ($K=1$) by setting the part size at $128 \times 128$, i.e., the size of the original character images. We denote this version as ``Ours as 1-shot'' in the later result. 
\end{itemize}
We compare these methods with our few-part-shot method with the part size of $32 \times 32$ and $K=13$ parts. 
This setup ($32 \times 32 \times 13$ total pixels) roughly equals one $128 \times 128$ image in terms of pixel count, allowing for an approximate comparison of information capacity. 

\subsubsection{Results}
Table~\ref{tab:table1} presents the quantitative comparison of our method (the bottom row) against other one-shot font generation approaches (FontDiffuser, FANnet + DM, and Ours as 1-shot). Our method significantly outperforms FANnet+DM in ContentFID and StyleFID, despite relying on far less information in the one-part scenario. This finding suggests that font generation can be achieved with minimal input and still achieve performance on par with simpler one-shot methods. In contrast, FontDiffuser exhibits higher performance than our model on all metrics, likely due to its carefully designed two-step training strategy that provides superior global coverage of font styles.\par

Nevertheless, our approach produces results comparable to or better than FontDiffuser in certain cases, which we examine in detail in the next section. More importantly, by allowing designers to supply only local shapes as opposed to a fully designed character, our few-part-shot framework offers significant flexibility in practical font creation. Although it currently cannot surpass state-of-the-art methods that use entire characters, the fact that it can come close underscores its potential. With the option to incorporate a few complete character references or additional global constraints, our method could be extended further, closing the gap with FontDiffuser while maintaining the benefits of local part-based design. Such a hybrid strategy opens up new avenues for efficient and creative font generation workflows.
\begin{table}[t]
    \centering
    \caption{Quantitative comparison with the existing methods for one-shot font generation. ``Ours as 1-shot'' is a one-shot version of the proposed method using the whole image of a single character (i.e., part size was $128\times 128$ and {$K=1$}). For the proposed method (Ours), the part size was $32\times 32$ and {$K=13$}. The column ``part'' shows whether one-part-shot or not. The column ``\#shot'' shows the number of shots for generation.}
    \begin{tabular}{l|c|c|c|c|c|c|c|c} \hline
         & part &\#s& L1$\downarrow$ & Local L1$\downarrow$& SSIM$\uparrow$ & MSSIM$\uparrow$ & StyleFID$\downarrow$ & ContentFID$\downarrow$ \\ \hline
         FontDiffuser &&1& \multicolumn{1}{|r|}{0.128} & \multicolumn{1}{|r|}{0.172} & \multicolumn{1}{|r|}{0.602} & \multicolumn{1}{|r|}{0.560} & \multicolumn{1}{|r|}{53.785} & \multicolumn{1}{|r}{4.026} \\
         FANnet+DM &&1& \multicolumn{1}{|r|}{0.164} & \multicolumn{1}{|r|}{0.227} & \multicolumn{1}{|r|}{0.565} & \multicolumn{1}{|r|}{0.445} & \multicolumn{1}{|r|}{51.139} & \multicolumn{1}{|r}{133.107} \\
         Ours as 1-shot &&1& \multicolumn{1}{|r|}{0.135} & \multicolumn{1}{|r|}{0.185} & \multicolumn{1}{|r|}{0.604} & \multicolumn{1}{|r|}{0.513} & \multicolumn{1}{|r|}{15.918} & \multicolumn{1}{|r}{18.276} \\ \hline
         Ours ($K=13$) &$\checkmark$&13& \multicolumn{1}{|r|}{0.171} & \multicolumn{1}{|r|}{0.228} & \multicolumn{1}{|r|}{0.540} & \multicolumn{1}{|r|}{0.426} & \multicolumn{1}{|r|}{15.296} & \multicolumn{1}{|r}{9.060} \\ \hline
    \end{tabular}
    \label{tab:table1}
\end{table}

\begin{figure}[t]
    \centering
    \includegraphics[width=0.8
    \linewidth]{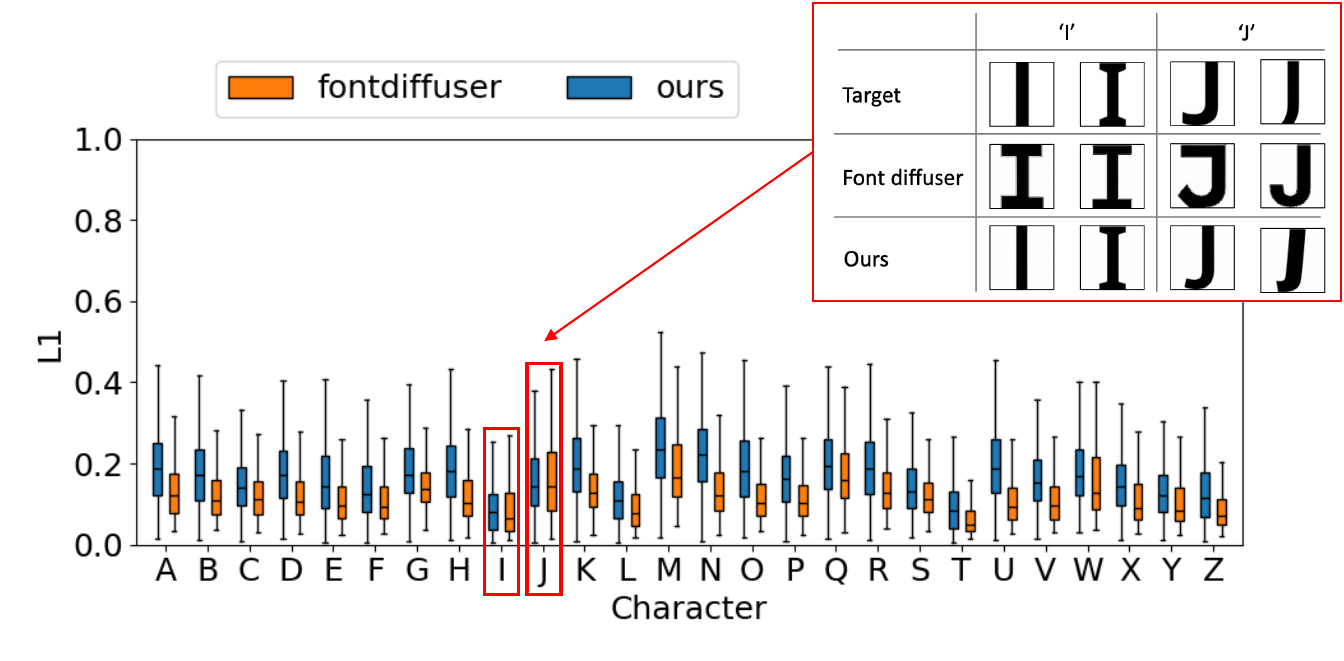}\\[-3mm]
    \caption{Class-wise evaluation of generated images with L1 distance. For the proposed method, the part size is set at $32\times{32}$ and $K=13$.}
    \label{fig:box-plot}
\end{figure}

\subsubsection{Class-wise comparison with FontDiffuser}

Fig.~\ref{fig:box-plot} shows a box-plot of L1 distances of FontDiffuser and ours for each character class, revealing variability in generation performance depending on the specific letter. The results for `I' and `J' (highlighted with a red box) show that our method performs comparably to FontDiffuser on these two characters. 
Qualitative analysis indicates a recurring trend, as illustrated in the top-right inset of the figure. 
FontDiffuser often introduces serifs when generating `I' and produces a thicker shape for `J,' especially in fonts where `I' and `J' are much narrower than other letters. In contrast, our method generates `I' and `J' with widths closer to those in the target fonts. Since our method leverages only local parts, it avoids certain global tendencies that might distort thin characters into a more uniform style. Moreover, `I' and `J' are very simple characters, composed of a single, mostly straight stroke; thus, their shapes can be reliably inferred from a small set of local design cues.

\section{Conclusion, Limitation, and Future Work}

In this paper, we challenged \emph{few-part-shot font generation}, a new extension of few-shot font generation that relies solely on localized parts rather than fully designed character images. By providing only small patches that capture stroke endings, corners, or decorative elements, our approach can generate fonts that largely reflect the intended style. Through a series of qualitative and quantitative experiments, we confirmed several capabilities and limitations:
\begin{itemize}
    \item Even a single, carefully selected part can convey enough local style cues to produce recognizable variations of an entire alphabet.
    \item Increasing either the part size or the number of parts further stabilizes and improves the quality of the generated fonts, though the benefits diminish beyond a certain point (i.e., diminishing marginal returns).
    \item A fundamental limitation arises in reproducing certain global characteristics (e.g., oblique angles or condensed shapes). Local parts alone cannot guarantee global consistency because they lack holistic information about overall proportions.
\end{itemize}
Despite the limitation, our few-part-shot approach offers a more flexible and efficient way to prototype fonts without requiring a fully finished reference character. Looking ahead, we plan to investigate hybrid methods that combine part-based inputs with a small number of complete-character references, hoping to capture both local and global structure more effectively. 
We also envision applying this method to more complex scripts (e.g., non-Latin alphabets) and exploring richer editing capabilities such as style interpolation or partial shape manipulation. 
By unifying local and global design cues in one framework, we aim to further bridge the gap between the creativity of human font designers and the power of modern generative models.

\begin{credits}
\subsubsection{\ackname} 
This work was supported by JSPS KAKENHI-JP25H01149 and JST CRONOS-JPMJCS24K4.
\end{credits}
%
%
%

\bibliographystyle{splncs04}
\bibliography{mybib}

\end{document}